%% file: latex/acl_latex.tex
\pdfoutput=1

\documentclass[11pt]{article}

\usepackage[preprint]{acl}

\usepackage{times}
\usepackage{latexsym}
\usepackage{todonotes}

\usepackage[T1]{fontenc}

\usepackage[utf8]{inputenc}

\usepackage{microtype}

\usepackage{inconsolata}

\usepackage{graphicx}
\usepackage{pgfplots}
\usepackage{geometry}
\usepackage{amsmath}
\usepackage{caption}   
\usepackage{cuted}

\usepackage{booktabs}
\usepackage{comment}
\pgfplotsset{compat=1.18}
\usepackage{xcolor}
\usepackage{fancyvrb}
\usepackage{listings}
\usepackage{hyperref}
\usepackage{multirow}
\usepackage{siunitx} 
\usepackage[export]{adjustbox}
\input{math_commands.tex}

\usepackage{hyperref}
\usepackage{url}
\usepackage{amsmath,amssymb,mathtools}
\usepackage{algorithm}
\usepackage{algpseudocode}
\usepackage{booktabs}
\usepackage{multirow}
\usepackage{makecell}
\usepackage{graphicx}  
\usepackage{siunitx}   
\usepackage{makecell}
\graphicspath{{figures/}} 
\usepackage{subcaption} 

\usepackage{mdframed}
\usepackage{needspace}
\usepackage{listings}

\usepackage{multicol}      

\usepackage{dblfloatfix}
\usepackage{lipsum}  
\definecolor{myred}{RGB}{234, 103, 96}
\definecolor{myblue}{RGB}{93, 169, 222}

\title{Activation-Space Personality Steering: Hybrid Layer Selection for Stable Trait Control in LLMs}

\author{
 \textbf{Pranav Bhandari\textsuperscript{1,2,*}}, 
  \textbf{Nicolas Fay\textsuperscript{3}},
    \textbf{Sanjeevan Selvaganapathy\textsuperscript{1,2}}, \\
  \textbf{Amitava Datta\textsuperscript{2}},
  \textbf{Usman Naseem\textsuperscript{4}},
 \textbf{Mehwish Nasim\textsuperscript{1,2,*}}
\\
 \textsuperscript{1}Network Analysis and Social Influence Modelling (NASIM) Lab \\
 \textsuperscript{2}School of Physics Maths and Computing, The University of Western Australia\\
 \textsuperscript{3}School of Psychological Science, The University of Western Australia \\
 \textsuperscript{4}School of Computing, Macquarie University
\\
  \small{
   \textbf{\textsuperscript{*}Correspondence:} {firstname.lastname@uwa.edu.au}
 }
}

\begin{document}
\maketitle
\begin{abstract}
Large Language Models exhibit implicit personalities in their generation, but reliably controlling or aligning these traits to meet specific needs remains an open challenge. The need for effective mechanisms for behavioural manipulation of the model during generation is a critical gap in the literature that needs to be fulfilled. Personality-aware LLMs hold a promising direction towards this objective. However, the relationship between these psychological constructs and their representations within LLMs remains underexplored and requires further investigation. Moreover, it is intriguing to understand and study the use of these representations to steer the models' behaviour. We propose a novel pipeline that extracts \textbf{hidden state activations} from transformer layers using the Big Five Personality Traits \emph{(Openness, Conscientiousness, Extraversion, Agreeableness and Neuroticism)},
which is a comprehensive and empirically validated framework to model human personality \textbf{ which applies low-rank subspace discovery} methods, and \textbf{identifies trait-specific optimal layers} across different model architectures for robust injection. The resulting personality-aligned directions are then operationalised through a flexible steering framework with dynamic layer selection, enabling precise control of trait expression in LLM outputs. Our findings reveal that personality traits occupy a low-rank shared subspace, and that these latent structures can be transformed into actionable mechanisms for effective steering through careful perturbations without impacting the fluency, variance and general capabilities, helping to bridge the gap between psychological theory and practical model alignment.
\end{abstract}

\section{Background}

\input{latex/sections/background}
\section{Related Work}
\input{latex/sections/related_work}
\section{Methodology}
\input{latex/sections/methodology}

\section{Evaluation}

\input{latex/sections/evaluation}

\section{Results}
\input{latex/sections/results}


\section{Conclusion}
\input{latex/sections/conclusion}

\section*{Acknowledgment}
\input{latex/sections/ack}

\section*{Limitations}
\input{latex/sections/limitations}

\section*{Ethical Considerations}
\input{latex/sections/ethical_considerations}

\bibliographystyle{acl_natbib}
\bibliography{custom}

\newpage
\appendix

\input{latex/sections/appendix}

\end{document}

%% file: math_commands.tex
\usepackage{amsmath,amsfonts,bm}

\def\eqref#1{equation~\ref{#1}}

\def\1{\bm{1}}

\DeclareMathAlphabet{\mathsfit}{\encodingdefault}{\sfdefault}{m}{sl}
\SetMathAlphabet{\mathsfit}{bold}{\encodingdefault}{\sfdefault}{bx}{n}



%% file: latex/sections/background.tex
Large Language Models (LLMs) are increasingly shaping human–computer interaction, influencing decisions in critical sectors such as healthcare, finance, and education \cite{chen2024survey,raza2025industrial}, yet their outputs often reflect uncontrolled or opaque behavioural tendencies. A plausible solution is \emph{steering} the models, where targeted interventions are applied to internal model activations at inference time without retraining to align the outputs with the desired attributes \cite{identifying_manipulating,steering_language_acteng,li_inference-time_2024,zhu2024personality}. Steering has been shown to effectively modulate properties such as sentiment, politeness, and toxicity in a lightweight and interpretable manner \cite{lai-etal-2024-style, naseem2026mechanistic}. While large-scale alignment methods such as reinforcement learning from human feedback (RLHF), reinforcement learning from AI feedback (RLAIF), proximal policy optimisation (PPO), or direct preference optimisation (DPO) have advanced alignment with human values, they are costly in terms of data, computation, and stability \cite{zhu2024personality, deng2024neuron}. These methods typically update model weights and often target narrow objectives such as truthfulness or honesty, leaving subtle descriptors such as personality traits underexplored.

Activation steering \cite{turner2023steering, li_inference-time_2024}  modifies a model's behaviour during inference by adjusting the residual stream of transformer layers, avoiding the need for retraining. While effective for simple, surface-level attributes, its use for complex traits like personality has not yet been thoroughly explored. 
Recent work has begun to investigate personality steering \cite{zhu2024personality, deng2024neuron}, drawing on psychological frameworks such as the Big Five \cite{john1999big} --- \emph{\textbf{O}penness, \textbf{C}onscientiousness, \textbf{E}xtraversion, \textbf{A}greeableness, \textbf{N}euroticism} (OCEAN) traits. These traits offer a natural lens for controllable generation that is directly relevant to personalisation, user alignment, and safety-sensitive applications. Despite this promise, several challenges persist: \emph{identifying stable trait directions}, \emph{deciding which layers to steer}, and \emph{verifying controlled shifts} without degrading the model's core competence.  

Most existing activation engineering methods show that adding a direction to the activation can steer behaviour, but they typically assume fixed layers \cite{turner2023steering} (e.g., layer 18 in LLaMA models) or narrow layer ranges for all prompts and concepts. Using default layers hinders effective steering for three reasons: \textbf{(i)} LLM architectures vary in depth, so ``middle layers'' do not translate consistently across models; \textbf{(ii)} different layers have different responsiveness, with sensitivity also varying across traits; and \textbf{(iii)} there is no principled method to balance layers, traits, and architectures. As a result, steering is often unreliable, non-reproducible, or misaligned with actual inference-time behaviour. To address this gap, we propose a two-stage method that combines \emph{static verification} (offline diagnostics to extract effective layer representations) with \emph{dynamic measurement} (runtime responsiveness to the given prompt). This hybrid approach yields a robust set of candidate injection layers for each trait, striking a balance between general reliability and prompt-specific adaptivity.

Our pipeline begins by extracting internal activations from a pretrained causal LLM using datasets annotated with high and low levels of each Big Five trait. These activations are standardised and used to derive trait-specific directions, which are then aggregated across layers to reflect the varying sensitivity of different parts of the model. To reduce redundancy and capture shared structure, we project these directions into a low-rank subspace, ensuring that personality steering remains compact and interpretable and encodes minimal noise. We then introduce a hybrid strategy for layer selection: static diagnostics identify generally reliable layers, while dynamic measurements on the current prompt capture runtime responsiveness. The two sources are combined into a candidate set for injection. Finally, during inference, the chosen trait direction is injected as a scaled perturbation into the residual stream of the selected layers via forward hooks, steering the model's generation toward the desired personality expression while preserving its core abilities. Our contributions are as follows: 

\begin{figure*}[t] 
  \centering
  \includegraphics[scale = 0.5,trim =20 0 0 0,clip]{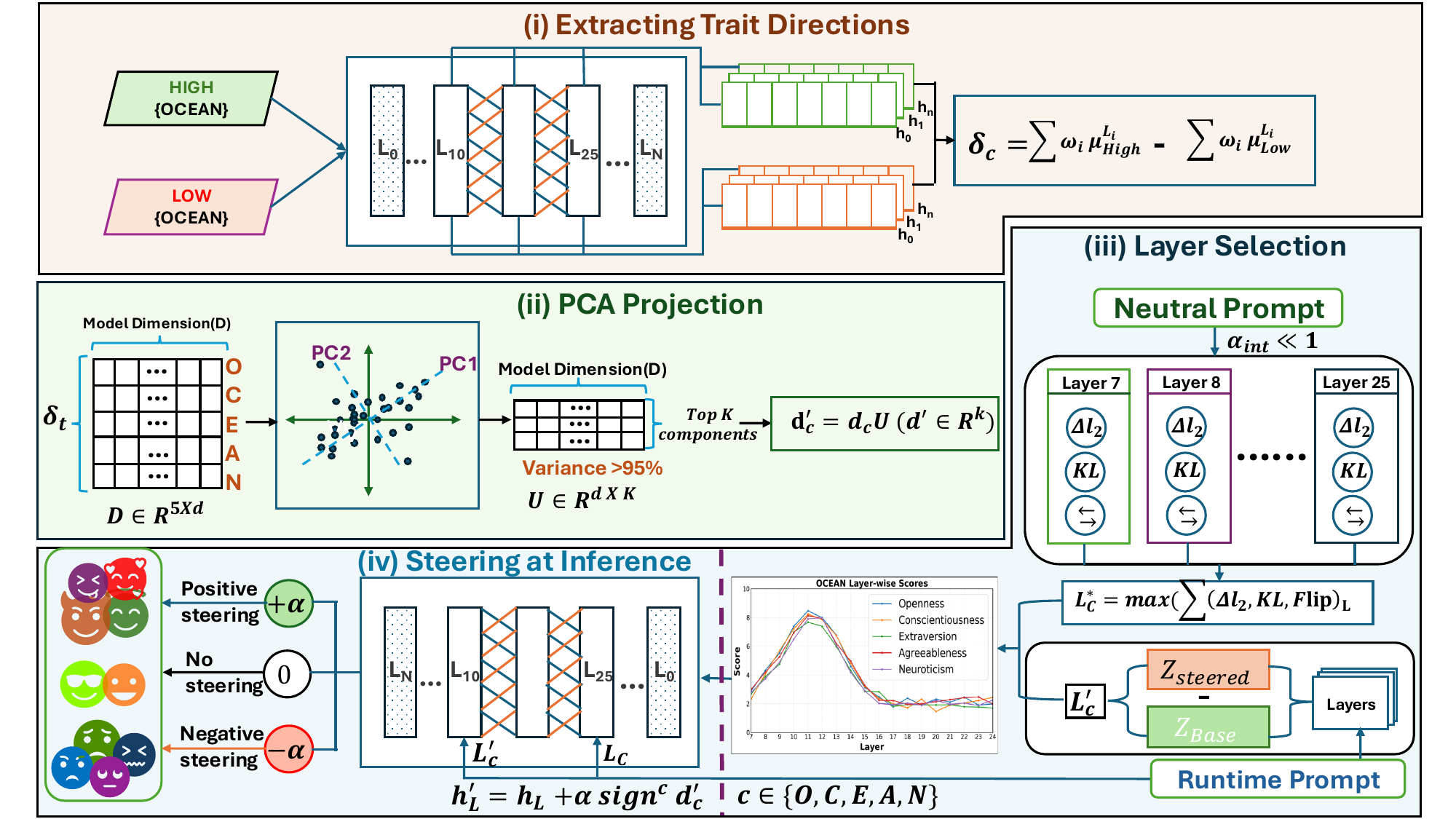}
 \caption{Diagram illustrating our methodology in four phases. \textbf{i. Extracting trait directions:} Trait-wise mean difference vectors are computed from \emph{High/Low} samples across layers. \textbf{ii. PCA projection:} The aggregated directions are projected onto a low-dimensional subspace to obtain orthogonal, normalised trait vectors. \textbf{iii. Layer selection:} The best injection layer is chosen using offline diagnostics (verified layer) combined with a prompt-specific dynamic layer for \emph{hybrid steering}. \textbf{iv. Steering at Inference: } The final intensity-scaled trait vectors are injected into the residual stream of the selected Layers guided by polarity to produce personality-aligned text. }
  \label{fig:methodology}
\end{figure*}


\begin{itemize}
    \item We propose an end-to-end pipeline operating across multiple levels: constructing contrastive, trait-labelled activation sets, deriving per-trait steering directions, selecting effective intervention layers, and injecting directions via forward hooks for each Big Five trait. 

    \item We stack trait directions and perform PCA/SVD to extract top-k orthonormal components used for steering. The resulting subspace projections (unit-normalised) reduce variance and noise, improve stability, and compress steering while retaining over 95$\%$ of inter-trait energy. 

    \item Instead of assuming a fixed middle layer, we introduce a hybrid method to locate trait-responsive layers. Diagnostic metrics ($\Delta \ell_2$, KL divergence, flip rate) identify static sensitivities, while dynamic $\Delta$-logit norms capture prompt-specific responsiveness. Merging static reliability with dynamic adaptivity yields context-aware, stable, and reproducible steering.
    
\end{itemize}

%% file: latex/sections/related_work.tex
\paragraph{Activation Engineering for Behaviour Control}

Previously studied activation engineering methods aim to modify internal model representations at inference time to steer behaviour without retraining. Techniques such as Contrastive Activation Addition \cite{rimsky_steering_2024} compute steering vectors by averaging the activation difference between positive and negative behaviour pairs, and are successful in modulating behaviours such as sycophancy and hallucination. Techniques like Representation Engineering \cite{zou_representation_2025} apply Linear Artificial Tomography and PCA on contrastive activations to control attributes such as honesty and power-seeking. Inference-Time Intervention (ITI) \cite{li_inference-time_2024} locates attention heads with distributional differences between true and false statements, improving TruthfulQA performance. While ITI relies on binary contrastive pairs and PAS on human-scored Likert data, both model traits independently without leveraging shared structure. However, a recent evaluation \cite{tan_analyzing_2025} highlights key limitations: high variance across inputs, sensitivity to prompt variations, and dependence on dataset biases over genuine concept modelling. Crucially, some concepts even prove to be `anti-steerable', steering in the reverse direction. Furthermore, studies such as \cite{silva_steering_2025} show that logit-lens based steering only succeeds on some model families and not others, and that function vectors work in only 20\% of model-task combinations without extensive hyperparameter search.

\paragraph{Personality Modelling in LLMs}

There have been numerous different approaches in historical research into (inducing) LLM personality. Prompting-based methods \cite{serapio-garcia_personality_2025, personallm} offer simple and lightweight controls, but suffer from surface-level trait expression and prompt sensitivity. Fine-tuning approaches demonstrate a much deeper personality integration: \cite{big5dataset} show that SFT/DPO outperform prompting while even correlating personality traits with reasoning performance, which aligns with psychological findings on how these traits (e.g., higher conscientiousness, higher agreeableness, lower extraversion, and lower neuroticism) correspond to an increase in cognitive performance in humans. While research in psychology establishes correlations between certain Big Five traits (e.g., openness and extraversion), existing computational methods ignore these relationships, modelling each trait in isolation. Other approaches \cite{sun_personality_2025} propose personality vectors via model merging, achieving continuous control over personality traits with transferability to multilingual and vision-language models. However, multi-trait composition is impeded by parameter interference, with control dropping when merging all five traits.

\paragraph{Layer Selection and Low-Rank Methods}

Identifying optimal intervention layers is critical to the efficacy of steering methods. Causal tracing \cite{meng_locating_2023} localises factual associations to middle MLP layers, and sentiment steering has also been shown to present the best performance when training and evaluating a layer near the middle of the model. CogSteer \cite{wang_cogsteer:_2025} proposes cognition-inspired layer buckets, achieving small improvements in benchmarks such as GLUE with only 3.1\% of parameters by only intervening in middle layers across multiple architectures. Yet, layer selection still remains task-dependent with narrow optimal ranges varying unpredictably across models; especially in models with larger architectures past those studied in the previous literature.
Low-rank subspace methods offer efficient alternatives to full-parameter interventions. Task Arithmetic \cite{ilharco_editing_2023} enables task composition through vector addition, with up to 10,000 times fewer parameters than fine-tuning. Orthogonal Subspace Learning enforces orthogonality constraints between task subspaces to prevent catastrophic forgetting. Gaussian Concept Subspaces \cite{zhao_beyond_2025} model concepts of physical objects/entities (e.g. animals, sports games) as distributions, rather than point estimates, for improved robustness. These methods together demonstrate that many behavioural controls occupy low-dimensional manifolds, motivating our approach to personality steering.


%% file: latex/sections/methodology.tex
We steer a pretrained causal LLM along interpretable personality factors --- the OCEAN traits,
by adding low-rank control vectors 
into the \emph{decoder residual stream} during decoding. Our method proceeds in four steps: 
(i) \emph{\textbf{E}stimate} layer-wise trait directions from a high/low labelled dataset and aggregate them into per-trait weighted directions, 
(ii) \emph{\textbf{L}earn} a low-dimensional subspace to capture shared personality structure 
(iii) \emph{\textbf{E}xtract} the offline best layer per trait (iv) \emph{\textbf{S}teer} at inference time through the hybrid layer selection method with a projected, intensity-scaled vector through forward hooks. 
Figure \ref{fig:methodology} presents our methodology. Detailed explanation is provided in the following sections. 
A key contribution of our approach is the \emph{hybrid layer selection} strategy (Section~\ref{sec:layer-selection}), 
which combines verified offline priors with dynamic, prompt-specific diagnostics. 
Unlike prior work that fixes a single injection layer \cite{turner2023steering,instruction_folloiwing_llms} or relies purely on 
static heuristics, our method balances stability (through verified layers) with adaptability 
(through runtime selection). This design makes steering more robust across traits, architectures, and prompts and allows us to test the method consistently across multiple evaluation scenarios 
(personality questionnaires, open-ended discourse, and general reasoning benchmarks).

\begin{figure*}[t] 
  \centering
  \includegraphics[width=\textwidth,trim=0 0 0 0,clip]{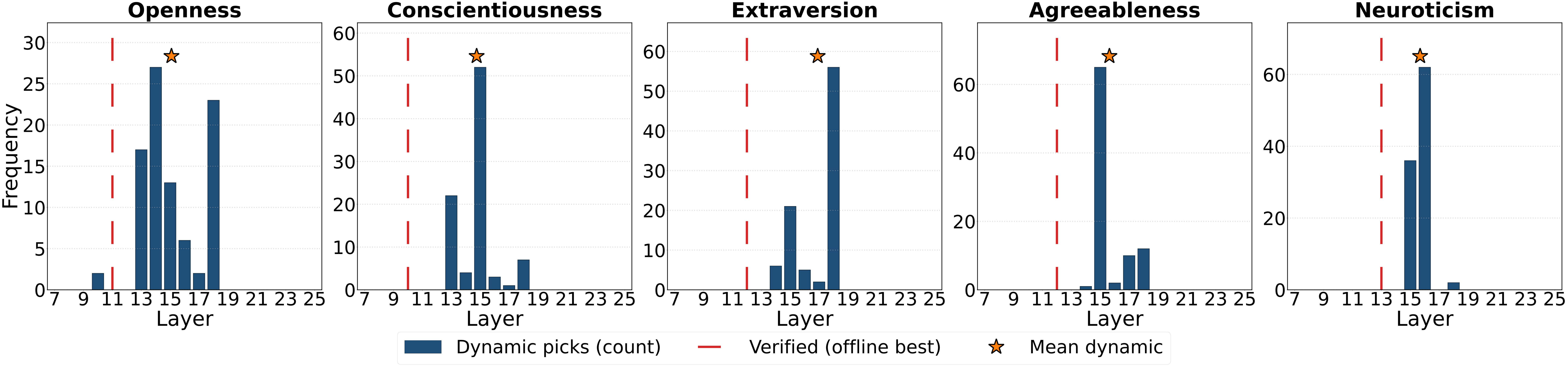}
  \caption{Diagrams representing the \emph{Hybrid Trait Selection} for each of the personality traits used in our methods for LLama-3-8B-Instruct model. Dotted lines represent the Offline Prior/Offline Best method that serves as a static layer selection. The layer vs. Frequency plot demonstrates the choice of layers and frequency during the steering process. $\star$ represents the mean of dynamic layers selected across all runs.  }
  \label{fig:hybrid_method_plot}
\end{figure*}

\subsection{Problem Setup}
Let $\mathcal{M}$ denote a causal LLM with parameters $\theta$ and vocabulary size $V$.
Given tokens $x_{1:T}$, the model produces residual states $\mathbf h_t\in\mathbb{R}^d$ and logits $\mathbf z_t\in\mathbb{R}^V$ with
\begin{align}
\mathbf{z}_t &= W\,\mathbf{h}_t \\
p(x_t\mid x_{<t};\theta) &= \mathrm{softmax}(\mathbf{z}_t) \\
p(x_{1:T}) &= \prod_{t=1}^{T}p(x_t\mid x_{<t};\theta)
\end{align}
We steer $\mathcal{M}$ at inference by adding a small, structured perturbation to the residual stream at selected decoder layers so that the generated text exhibits a target Big-Five trait $c\in\mathcal{C}=\{\text{O},\text{C},\text{E},\text{A},\text{N}\}$ with controllable intensity, \emph{without} changing~$\theta$.

\subsection{Activation Extraction + Standardisation}
For each trait $c$, we use a subset of the Big-5-Chat\footnote{\href{wenkai-li/big5_chat}{Huggingface-BIG5-Chat}} dataset \cite{big5dataset} that contains $20,000$ instances partitioned into \emph{high}/\emph{low} labels (5,000 for each high and low for each trait). 
For a candidate layer index $L$, we extract the \emph{last non-pad} residual state per sequence $i$,
$\mathbf h^{(i)}_{L}\in\mathbb{R}^d$.
We then \emph{jointly} standardise high and low activations using a shared mean/variance for each $(L,c)$. Let the class mean be $\boldsymbol{\mu}^{(c)}_{L}(v)=\frac{1}{N_v}\sum_{i:\,y_i=v}\mathbf{h}^{(i)}_{L}$ for $v\in\{\mathrm{high},\mathrm{low}\}$, where $y_i$ denotes the high/low label. The normalised mean-difference (trait) direction: 
\begin{equation}
\mathbf{d}^{(c)}_{L}
= \frac{\boldsymbol{\mu}^{(c)}_{L,\mathrm{high}}-\boldsymbol{\mu}^{(c)}_{L,\mathrm{low}}}
      {\bigl\|\boldsymbol{\mu}^{(c)}_{L,\mathrm{high}}-\boldsymbol{\mu}^{(c)}_{L,\mathrm{low}}\bigr\|_2}
\in\mathbb{R}^d
\end{equation}
Because layers vary in discriminative power, we learn non-negative \emph{trait-specific} weights $\{w^{(c)}_L\}$ (summing to 1) that emphasise layers that separate high vs.\ low for trait $c$. The \emph{aggregated} direction is
$\mathbf d^{(c)}=\sum_{L\in\mathcal{L}} w^{(c)}_L\,\mathbf d^{(c)}_{L}$
yielding one robust per-trait direction that integrates evidence across layers.

\subsection{Low-Rank Personality Subspace}
We stack the aggregated per-trait directions $\{\mathbf d^{(c)}\}$ and fit a rank-$k$ PCA basis $U_k\in\mathbb{R}^{d\times k}$ (orthonormal columns).
Any trait vector is projected and renormalised as
\begin{align}
\tilde{\mathbf{d}}^{(c)} &= U_kU_k^\top\mathbf{d}^{(c)} \\
\widehat{\mathbf{d}}^{(c)} &= \frac{\tilde{\mathbf{d}}^{(c)}}{\|\tilde{\mathbf{d}}^{(c)}\|_2}
\end{align}
At inference, the steerer uses $\widehat{\mathbf d}^{(c)}$; both $U_k$ and the per-trait aggregated vectors are stored as artefacts along with the learnt layer weights.

\subsection{Layer Selection: Verified + Dynamic Hybrid Strategy}
\label{sec:layer-selection}
The selection of the injection layer is critical, as different layers vary in their response to steering. Some are highly sensitive, while others show little effect; prior work often fixed mid-layers (e.g., 18), but this ignores trait and prompt specific variation. We adopt a two-stage strategy: (i) pre-computed verified layers as trait-specific priors, and (ii) lightweight dynamic evidence per prompt.  

\paragraph{(1) Offline Prior (Best Layer Per Trait).}
For each trait $c$, we identify a single best layer $L_c^\star$ using neutral probe prompts (short generic instructions without trait words) and measure small-signal steering at the next token position. Let $\mathbf p_0$ be the baseline distribution and $\mathbf p^{(c)}_{1,L}$ the distribution after a tiny steer ($\alpha_{\text{probe}}\!\ll\!1$) at layer $L$. Three diagnostics are computed:
\begin{gather}
\Delta\ell_2(L,c) =\lVert \mathbf p^{(c)}_{1,L}-\mathbf p_0\rVert_2\\
\mathrm{KL}(L,c) =\sum_v p_0(v)\log\frac{p_0(v)}{p^{(c)}_{1,L}(v)}\\
\phi(L,c)=\mathbb{E}\!\big[\mathbf 1(\arg\max \mathbf p_0 \neq \arg\max \mathbf p^{(c)}_{1,L})\big]
\end{gather}
These capture \emph{raw sensitivity}($\Delta\ell_2$), \emph{semantic shift in high-probability tokens}(KL), and \emph{categorical flips}($\phi$). We combine them with
\begin{equation}
S(L,c)=\lambda_{\ell_2}\,\Delta\ell_2+\lambda_{\mathrm{KL}}\,\mathrm{KL}+\lambda_{\mathrm{flip}}\,\phi
\end{equation}
and select $L_c^\star=\arg\max_L S(L,c)$. The weights $\lambda$ are fixed once to balance the magnitudes, not tuned per trait, and the resulting best layer serves as a stable offline prior.

\paragraph{(2) Dynamic Runtime Selection.} Since prompts activate layers differently, we add a simple runtime diagnostic. For a prompt $p$, the per-layer shift \begin{equation} \nu(L,p) = \|\mathbf{z}^{\text{steered}}_{L}(p)-\mathbf{z}^{\text{base}}(p)\|_2 \end{equation} is computed, and the layer with highest $\nu(L,p)$ is selected as the dynamic candidate $\mathcal R(p,c)$. This adapts layer choice to the input context.

\paragraph{(3) Hybrid Combination.}  
At inference we steer jointly at the verified prior $\mathcal V_c=\{L^*_c\}$ and the dynamic layer $\mathcal R(p,c)$.  
To balance stability and adaptability, we fix mixture weights $(0.8,0.2)$: the heavier verified weight reflects its offline reliability, while the lighter dynamic weight injects context sensitivity.  
This heuristic, chosen for both reproducibility and practical robustness, ensures that the method is not biased by spurious runtime spikes while remaining responsive to the prompt. Figure \ref{fig:hybrid_method_plot} shows runtime results from our \emph{Hybrid Trait Selection} method.



\paragraph{(4) Intensity Parameter($\alpha$) Selection.}
\label{para:alpha_selection}

The steering intensity $\alpha$ determines how strongly the trait vector affects the model's residual activations where too small yields negligible change and too large harms coherence. To ensure interpretability and stability, $\alpha$ was empirically tuned through a controlled sweep, tracking both \emph{average trait} and \emph{fluency scores}.
The goal was to identify the highest $\alpha$ that maintained the quality of natural language, specifically ensuring that the average fluency score in either direction (positive or negative steering) did not drop below the threshold value of $3.5$ on the scale of 1-5. 
We use \emph{absolute scaling}, meaning that the steering vector is applied with the same magnitude $\alpha$ across prompts, independent of the layer norm or hidden-state statistics. This provides consistent intensity across traits and avoids prompt-dependent variability. We report the values of $\alpha$ used in our experiments that correspond to the respective results. 
\subsection{Polarity Calibration}
The direction of each trait $\widehat{\mathbf{d}}^{(c)}$ may initially point toward or against the intended semantic effect (e.g., high vs.\ low trait). To resolve this ambiguity, we apply small steering ($\alpha << 1$) in both directions ($+\widehat{\mathbf{d}}^{(c)}$ and $-\widehat{\mathbf{d}}^{(c)}$) on a neutral calibration set $\mathcal{P}_{\text{cal}}$ and choose the sign $\mathrm{sign}^{(c)}=\arg\max_{s\in\{\pm1\}}\mathbb{E}_{p\in\mathcal{P}_{\text{cal}}}[\mathrm{KL}(p_0\,\|\,p_{1,s}^{(c)})]$ that induces the stronger and more consistent divergence from the baseline distribution. This ensures that the positive steering direction aligns with the direction that most meaningfully shifts the model's next-token probabilities. Finally, the assigned polarity is semantically verified using a small set of labelled high/low trait prompts; if the positive direction produces responses judged as more aligned with the high-trait description (e.g., talkative for Extraversion, organised for Conscientiousness), it is retained; otherwise, the sign is flipped. 

\subsection{Steering at Inference (Forward Hooks)}
At each decode step, for trait $c$ and the chosen layers $\mathcal{L}$ we add
\begin{align}
\Delta^{(c)}(\alpha) &= \alpha\,\mathrm{sign}^{(c)}\,\widehat{\mathbf{d}}^{(c)} \\
\mathbf{h}'_{L} &= \mathbf{h}_{L}+\Delta^{(c)}(\alpha)
\end{align}
to the residual stream of decoder block(s). With LM head $W$, the next-token distribution is
$\mathbf z'_{t}=W\,\mathbf h'_{L},
p'(x_t)=\mathrm{softmax}(\mathbf z'_{t})$

A single \emph{global steer gain} $g=8.0$ controls the maximum effective steering strength. Per-trait $\alpha$ values are scaled relative to this global gain, making $g$ effectively a ``volume knob'' for steering intensity. This design choice keeps our comparisons interpretable: \textbf{absolute scaling} guarantees reproducibility across datasets, while the \textbf{global gain} provides a uniform upper bound on intervention strength.

\begin{table*}[t]
  \centering
  \small
  \caption{Trait and fluency scores with variance (variance computed across \textbf{High/Base/Low} \emph{trait} scores per trait; arrows mark High ($\uparrow$) and Low ($\downarrow$)).}
  \label{tab:trait-fluency-var}
  \renewcommand{\arraystretch}{1.15}
  \setlength{\tabcolsep}{4pt}
  \resizebox{\textwidth}{!}{%
  \begin{tabular}{ll *{15}{c}}
    \toprule
    \multicolumn{2}{c}{} &
    \multicolumn{3}{c}{\textbf{Openness}} &
    \multicolumn{3}{c}{\textbf{Conscientiousness}} &
    \multicolumn{3}{c}{\textbf{Extraversion}} &
    \multicolumn{3}{c}{\textbf{Agreeableness}} &
    \multicolumn{3}{c}{\textbf{Neuroticism}} \\
    \cmidrule(lr){3-5}\cmidrule(lr){6-8}\cmidrule(lr){9-11}\cmidrule(lr){12-14}\cmidrule(lr){15-17}
    \multicolumn{2}{c}{} &
    \makecell{\textit{Trait}\\\textit{score}} & \makecell{\textit{Fluency}\\\textit{score}} & \textit{Var} &
    \makecell{\textit{Trait}\\\textit{score}} & \makecell{\textit{Fluency}\\\textit{score}} & \textit{Var} &
    \makecell{\textit{Trait}\\\textit{score}} & \makecell{\textit{Fluency}\\\textit{score}} & \textit{Var} &
    \makecell{\textit{Trait}\\\textit{score}} & \makecell{\textit{Fluency}\\\textit{score}} & \textit{Var} &
    \makecell{\textit{Trait}\\\textit{score}} & \makecell{\textit{Fluency}\\\textit{score}} & \textit{Var} \\
    \midrule
    \multirow{3}{*}{\makecell[l]{\textbf{Llama-3-8B-Instruct}
    }}
      & \textbf{High}
        & \(4.0\,\uparrow\) & 5.0 & \phantom{0}0.20
        & \(4.2\,\uparrow\) & 4.8 & \phantom{0}0.76
        & \(4.5\,\uparrow\) & 4.9 & \phantom{0}0.44
        & \(4.5\,\uparrow\) & 4.7 & \phantom{0}0.46
        & \(4.0\,\uparrow\) & 3.4 & \phantom{0}1.10  \\
      & \textbf{Base}
        & 3.6 & 4.7 & \phantom{0}0.84
        & 3.5 & 4.7 & \phantom{0}0.46
        & 2.7 & 5.0 & \phantom{0}1.52
        & 4.2 & 4.7 & \phantom{0}1.06
        & 1.6 & 4.3 & \phantom{0}1.20 \\
      & \textbf{Low}
        & \(2.8\,\downarrow\) & 3.8 & \phantom{0}0.96
        & \(1.4\,\downarrow\) & 3.4 & \phantom{0}0.61
        & \(1.5\,\downarrow\) & 4.3 & \phantom{0}0.89
        & \(1.3\,\downarrow\) & 3.9 & \phantom{0}0.44
        & \(1.0\,\downarrow\) & 5.0 & \phantom{0}0.3 \\
    \midrule
    \multirow{3}{*}{\makecell[l]{\textbf{Ministral-8B-Instruct}
    }}
      & \textbf{High}
        & \(3.9\,\uparrow\) & 5.0 & \phantom{0}0.00
        & \(3.2\,\uparrow\) & 4.7 & \phantom{0}0.98
        & \(4.1\,\uparrow\) & 4.3 & \phantom{0}0.85
        & \(4.6\,\uparrow\) & 4.0 & \phantom{0}0.60
        & \(2\,\uparrow\) & 4.1 & \phantom{0}1.40 \\
      & \textbf{Base}
        & 3.1 & 5.0 & \phantom{0}0.00
        & 2.8 & 4.6 & \phantom{0}0.98
        & 3.2 & 4.5 & \phantom{0}1.40
        & 3.8 & 4.2 & \phantom{0}1.40
        & 1.0 & 4.8 & \phantom{0}0.00 \\
      & \textbf{Low}
        & \(1.5\,\downarrow\) & 3.4 & \phantom{0}0.00
        & \(1.2\,\downarrow\) & 3.5 & \phantom{0}0.17
        & \(1.3\,\downarrow\) & 3.4 & \phantom{0}0.18
        & \(1.3\,\downarrow\) & 3.7 & \phantom{0}0.80
        & \(1.0\,\downarrow\) & 5.0 & \phantom{0}1.30 \\
    \midrule
    \multirow{3}{*}{\makecell[l]{\textbf{Ministral-24B-Instruct}}}
      & \textbf{High}
        & \(4.5\,\uparrow\) & 4.4 & \phantom{0}0.25
        & \(3.1\,\uparrow\) & 4.8 & \phantom{0}1.23
        & \(4.6\,\uparrow\) & 4.2 & \phantom{0}0.48
        & \(4.2\,\uparrow\) & 4.3 & \phantom{0}1.30
        & \(3.9\,\uparrow\) & 3.8 & \phantom{0}0.61 \\
      & \textbf{Base}
        & 3.7 & 5.0 & \phantom{0}0.41
        & 2.5 & 4.5 & \phantom{0}0.69
        & 3.1 & 4.6 & \phantom{0}1.11
        & 3.8 & 4.4 & \phantom{0}1.95
        & 1.6 & 4.5 & \phantom{0}1.20 \\
      & \textbf{Low}
        & \(2.7\,\downarrow\) & 4.9 & \phantom{0}0.25
        & \(1.2\,\downarrow\) & 3.4 & \phantom{0}0.22
        & \(1.1\,\downarrow\) & 4.3 & \phantom{0}0.48
        & \(1.6\,\downarrow\) & 3.9 & \phantom{0}0.67
        & \(1.1\,\downarrow\) & 4.5 & \phantom{0}0.11 \\
    \midrule
    \multirow{3}{*}{\makecell[l]{\textbf{Qwen-14B-Instruct}}}
      & \textbf{High}
        & \(3.7\,\uparrow\) & 4.3 & \phantom{0}0.16
        & \(3.9\,\uparrow\) & 4.8 & \phantom{0}0.53
        & \(4.0\,\uparrow\) & 4.7 & \phantom{0}0.75
        & \(3.7\) & 4.6 & \phantom{0}1.40
        & \(3.7\,\uparrow\) & 4.5 & \phantom{0}0.20 \\
      & \textbf{Base}
         & 3.5 & 5.0 & \phantom{0}0.60
        & 3.2 & 4.7 & \phantom{0}0.66
        & 1.8 & 4.4 & \phantom{0}0.68
        & 3.9 & 4.8 & \phantom{0}0.80
        & 1.3 & 5.0 & \phantom{0}0.06 \\
      & \textbf{Low}
        & \(1.8\,\downarrow\) & 5.0 & \phantom{0}0.16
        & \(1.0\,\downarrow\) & 4.3 & \phantom{0}0.75
        & \(1.4\,\downarrow\) & 4.3 & \phantom{0}0.10
        & \(1.8\,\downarrow\) & 4.7 & \phantom{0}1.10
        & \(1.0\,\downarrow\) & 5.0 & \phantom{0}0.10\\

     \midrule
    \multirow{3}{*}{\makecell[l]{\textbf{Gemmma-3-4b-IT}}}
      & \textbf{High}
        & \(4.2\,\uparrow\) & 3.6 & \phantom{0}0.20
        & \(3.4\,\uparrow\) & 3.8 & \phantom{0}0.40
        & \(4.8\,\uparrow\) & 3.2 & \phantom{0}0.40
        & \(4.3\, \uparrow\) & 3.6 & \phantom{0}.60
        & \(4.2\,\uparrow\) & 3.5 & \phantom{0}0.00 \\
      & \textbf{Base}
         & 3.7 & 4.0 & \phantom{0}0.41
        & 2.4 & 3.9 & \phantom{0}1.01
        & 2.8 & 3.9 & \phantom{0}0.40
        & 3.4 & 3.8 & \phantom{0}1.11
        & 2.6 & 4.0 & \phantom{0}0.98 \\
      & \textbf{Low}
        & \(1.4\,\downarrow\) & 3.5 & \phantom{0}0.65
        & \(1.1\,\downarrow\) & 3.1 & \phantom{0}0.00
        & \(1.3\,\downarrow\) & 3.6 & \phantom{0}0.23
        & \(1.6\,\downarrow\) & 3.7 & \phantom{0}0.89
        & \(1.5\,\downarrow\) & 3.4 & \phantom{0}0.75\\
    \bottomrule
  \end{tabular}%
  }
\end{table*}

%% file: latex/sections/evaluation.tex
We evaluate our steering methods across multiple models in three configurations: (1) base (no perturbation), (2) positively steered, and (3) negatively steered. Steering efficiency is compared against the base model across metrics assessing generation quality and general ability retention.

\subsection{Test Through Generation}

\paragraph{Personality Trait Questionnaires:}  
Personality questionnaires are a standard method for assessing personality in humans \cite{john1999big} and have also been widely applied to LLMs \cite{bhandari2025evaluating,sorokovikova2024llms}. We use them for evaluation in three settings: the default (no steering), positive steering, and negative steering. No use of any kind of customisation in the prompts was ensured during generation. However, questionnaires alone often underestimate the behavioural, cognitive, and emotional nuances \cite{suhr2023challenging}. To overcome this, we complement them with generation-based methods, ensuring that personality is captured not only through scores but also through demonstrated behaviour, thereby aligning with our broader generation-based testing. Questionnaires from the Big Five Inventory \cite{fossati2011big} are used in an interview style similar to \cite{wang2024incharacterevaluatingpersonalityfidelity}.

\paragraph{Personality Benchmark dataset:}One of the most reliable and commonly used methods in the literature \cite{personallm, frisch-giulianelli-2024-llm} is the generation of multiple scenarios. Following the evaluation protocols of \cite{deng2024neuron}, we use questions constructed from the SocialIQA \cite{sap2019socialiqa} dataset to create situational queries for generation tasks. These questions are provided to the base model as well as the positively and negatively steered models. The generated outputs are then validated in two ways, using GPT-based evaluation. We report the average Trait and Fluency scores from the GPT-based evaluation, along with the variance scores.

\subsection{General Capability Retention}
 
Excessive perturbations or poorly calibrated activation shifts can cause the model to degrade in its broader reasoning and problem-solving capabilities. Steering a language model toward the extreme positive or negative directions of a personality trait may be desirable, albeit preserving its general ability remains equally essential.
To validate this, we evaluate the steered models on two challenging and widely recognised benchmarks: MMLU \cite{wang2024mmlu}, which specialises in assessing knowledge and reasoning across diverse academic and professional domains, and ARC-Challenge \cite{clark2018think}, a benchmark designed to test complex reasoning and problem solving under more difficult settings. For the MMLU\footnote{\href{https://huggingface.co/datasets/cais/mmlu}{Huggingface MMLU}} dataset we tested across the validation sets of 11 different topics, and for the ARC-Challenge \footnote{\href{https://huggingface.co/datasets/allenai/ai2_arc}{Huggingface ARC-Challenge}}, we tested across 500 different questions with the same steering settings as generation.

%% file: latex/sections/results.tex

We used \texttt{LLaMA-3-8B-Instruct} as the primary base model to test steering,
and extended the evaluation to include broader families and parametric sizes: \emph{mistralai/Ministral-8B-Instruct-2410, mistralai/Mistral-Small-24B-Instruct-2501, Qwen/Qwen2.5-14B-Instruct, google/gemma-3-4b-it} to test the generalisation of our method. Across all experiments, the decoding parameters were fixed as
\texttt{temperature}~$=0.4$, \texttt{top\_p}~$=0.95$,
\texttt{top\_k}~$=50$, and \texttt{repetition\_penalty}~$=1.1$.
A moderate temperature of~$0.4$ was chosen instead of~$0$ to maintain slight lexical variability and prevent deterministic collapse of the distribution, which can exaggerate the apparent steering effects. $\alpha$ values were selected empirically as mentioned in Section \ref{para:alpha_selection} considering the fluency ($\geq 3.5$). Values for $\alpha$ ranged from $4 - 12$. 

\subsection{Low-Rank Subspace Assumption}

Table~\ref{tab:pc-variance-models} provides empirical justification for the low-rank formulation used in our method. For each model, we apply PCA to the matrix formed by the aggregated per-trait direction vectors (Openness, Conscientiousness, Extraversion, Agreeableness, Neuroticism) in model activation space. Across all models, the top three principal components explain more than 90\% of the total variance, indicating that personality-related activation differences concentrate in a shared, low-dimensional subspace.

This observation motivates the use of a low-rank basis as a compact and stable representation of personality information. Rather than enforcing low-rank structure, PCA reveals that the dominant variation across trait directions is intrinsically low-dimensional and consistent across model sizes and architectures. In our method, this basis is used as a projection space to retain the dominant trait-related signal while suppressing unstable or noisy components, improving the robustness of downstream steering. 

\begin{table}[h!]
\small
\centering
\caption{Explained variance of the top principal components (PC1--PC3) of the personality subspace across different LLMs. \textbf{Cumulative} denotes PC1+PC2+PC3.}
\label{tab:pc-variance-models}

\begin{tabular}{lcccc}
\toprule
\textbf{Model} & \textbf{PC1} & \textbf{PC2} & \textbf{PC3} & \textbf{$\sum_{i=1}^{3} \text{PC}_i$} \\
\midrule
LLaMA-8B  & 53.13 & 27.54 & 15.64 & 96.31 \\
Ministral-8B & 53.77 & 29.06 & 13.51 & 96.35 \\
Ministral-24B & 55.59 & 26.97 & 13.34 & 95.91 \\
Qwen-14B& 50.18 & 26.00 & 17.19 & 93.37 \\
\bottomrule
\label{tab:pca_reference}
\end{tabular}
\end{table}

\begin{figure}[t]
  \centering
  \begin{subfigure}[b]{0.48\textwidth}
    \includegraphics[width=\textwidth]{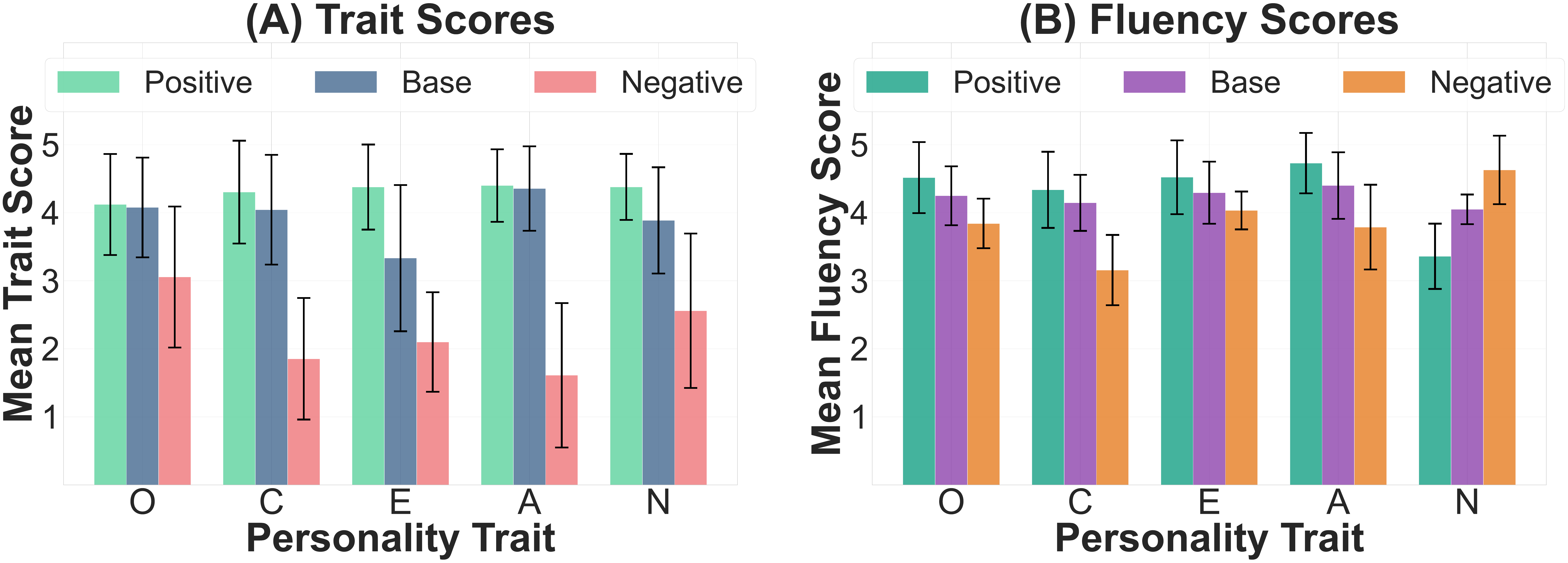}
    \caption{LLama-3-8B-Instruct}
    \label{fig:figure1}
  \end{subfigure}
  \hfill
  \begin{subfigure}[b]{0.48\textwidth}
    \includegraphics[width=\textwidth]{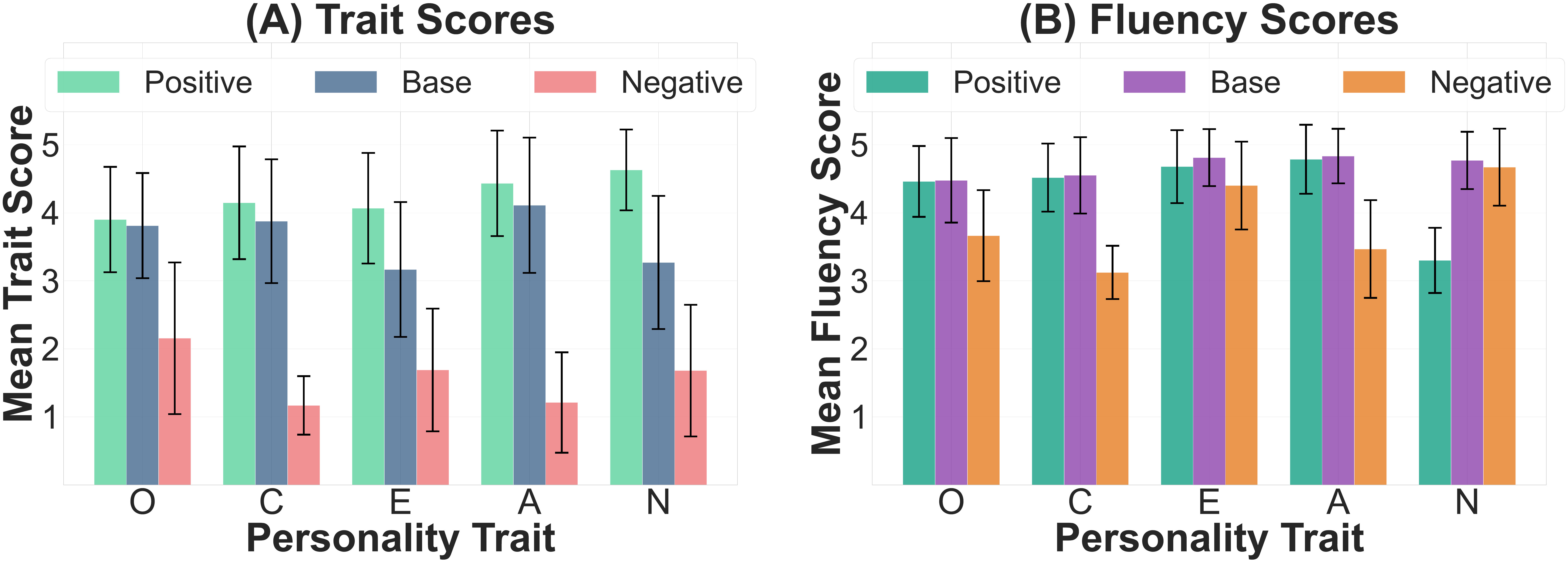}
    \caption{Ministral-8B-Instruct}
    \label{fig:figure2}
  \end{subfigure}
  \caption{Bar charts representing the \textbf{High/Base/Low} Traits and Fluency Scores with their corresponding variance scores.}
  \label{fig:situational_generations}
\end{figure}
 
\subsection{Test Through Generation}

\paragraph{Personality Trait questionnaires:} Table \ref{tab:trait-fluency-var} presents the average trait and fluency scores, along with the variance of the trait scores for five different models. For each of the five models, the effectiveness of steering is prominent in both positive and negative directions for all traits, thereby proving that our method is generalizable across models of different family and parametric sizes. For \textbf{LLaMA} model, the \textbf{trait separation} $\boldsymbol{\Delta}$ ranges from $\mathbf{1.2}$ to $\mathbf{3.2}$ with an \textbf{average of $\mathbf{2.64}$}. Other steering methods, such as \textit{prompt-based}, \textit{SFT}, and \textit{DPO} from \cite{big5dataset}, report averages of \textbf{2.6}, \textbf{2.7}, and \textbf{2.9}, respectively, based on the \textbf{direct scoring method} as opposed to the \textbf{generative method} that we use.
 Although the results are similar, our approach has key advantages: steering is applied at inference via low-rank control vectors without modifying the model weights $\theta$, which avoids overfitting or degraded fluency that often occurs in SFT/DPO \cite{easysteer, zhu2024personality}. In addition, our \emph{verified+dynamic} layer selection ensures that trait effects are both strong and context-sensitive, reducing brittleness across prompts. Finally, unlike most prompt or fine-tuning methods that require separate conditioning for high and low directions, our method naturally supports bidirectional steering within the same framework.

Beyond trait scores, monitoring fluency scores is crucial to ensure steering does not impair the model's general abilities. Prior studies note fluency degradation at extreme steering levels \cite{easysteer,steering_language_acteng}, but our method maintains stable fluency relative to the base model for either direction.
For Openness, Conscientiousness, and Neuroticism, the results show that steering models toward their positive trait even increases the fluency scores. Finally, the stability of the scores drastically improved for the majority of the traits such as Openness (0.84 $\rightarrow$ 0.2), Extraversion (1.52 $\rightarrow$ 0.89, 0.44), Agreeableness (1.06 $\rightarrow$ 0.46, 0.44), and Neuroticism (1.2 $\rightarrow$ 0.3, 1.1). This means that our approach not only preserves fluency under steering but also reduces variance, making the effects of steering more consistent and reliable across multiple runs, unlike prior methods, where high variance often indicated instability or degraded performance.

\paragraph{Personality Benchmark Dataset:} Figure \ref{fig:situational_generations} and \ref{fig:gemma_vis} represents the statistics for the situation based questions. Our steering method produces strong and consistent trait separations across both models, not only with the personality traits questionnaires but also in the benchmarking dataset. For \textbf{LLaMA-3-8B-Instruct}, positive vs.\ negative steering yields an average separation of 
$\Delta \approx 2.1$ on the 1--5 scale, with fluency largely preserved ($>4.0$). For \textbf{Ministral-3-8B-Instruct}, separations are even stronger 
($\Delta \approx 2.7$--$3.2$), though at the cost of higher variance and occasional fluency drops 
(e.g., Neuroticism-pos $=3.3$). 
Across traits, positive steering stabilises outputs (lower variance, higher fluency), while negative 
steering increases variability and slightly reduces fluency. In conclusion, LLaMA shows better \emph{fluency stability}, while Ministral exhibits stronger 
\emph{trait controllability}, highlighting a trade-off between linguistic stability and steering sensitivity.

\subsection{General Capability Retention}
Table \ref{tab:mmlu-arc-alpha4-final} and Table \ref{tab:mmlu-arc-alpha4-final_mistral}(\ref{sec:app}) report accuracy on MMLU and ARC-Challenge under different steering directions for LLaMA-3-8B-Instruct and Ministral-8B-Instruct, respectively. The results show that the overall performance remains 
stable around the base level, with only small fluctuations across traits. In particular, MMLU accuracy is well-preserved, while ARC shows minor variation depending on the direction of the trait. No catastrophic degradation occurs, indicating that personality steering preserves the model's reasoning and knowledge abilities.

\begin{table}[t]
  \centering
  \footnotesize
  \renewcommand{\arraystretch}{1}
  \setlength{\tabcolsep}{4pt}
  \resizebox{0.8\columnwidth}{!}{%
    \begin{tabular}{
      l
      S[table-format=2.2]  
      S[table-format=+2.2] 
      S[table-format=3.0]  
      S[table-format=+2.0] 
    }
    \toprule
    \textbf{Condition} & {\textbf{MMLU}} & {\textbf{$\Delta$ (pp)}} & {\textbf{ARC}} & {\textbf{$\Delta$ (pp)}} \\
    \midrule
    Base   & 69.27 & +0.00 & 84 & +0 \\
    \midrule
    Op +   & 69.72 & +0.45 & 82 & -2 \\
    Op -   & 67.06 & -2.21 & 78 & -6 \\
    Con +  & 68.43 & -0.84 & 80 & -4 \\
    Con -  & 68.35 & -0.92 & 84 & +0 \\
    Ext +  & 67.43 & -1.84 & 78 & -6 \\
    Ext -  & 68.35 & -0.92 & 82 & -2 \\
    Agr +  & 68.35 & -0.92 & 84 & +0 \\
    Agr -  & 69.72 & +0.45 & 84 & +0 \\
    Neu +  & 67.43 & -1.84 & 86 & +2 \\
    Neu -  & 69.27 & +0.00 & 80 & -4 \\
    \bottomrule
    \end{tabular}%
  }
  \caption{Results for \textbf{LLaMA-3-8B-Instruct} on MMLU and ARC-Challenge accuracy (\%) at $\alpha=4$.
  $\Delta$ columns show change vs.\ Base in percentage points (pp).}
  \label{tab:mmlu-arc-alpha4-final}
\end{table}

\subsection{Ablation Studies}
\begin{figure}[t] 
  \centering\fbox{\includegraphics[width=0.96\columnwidth]{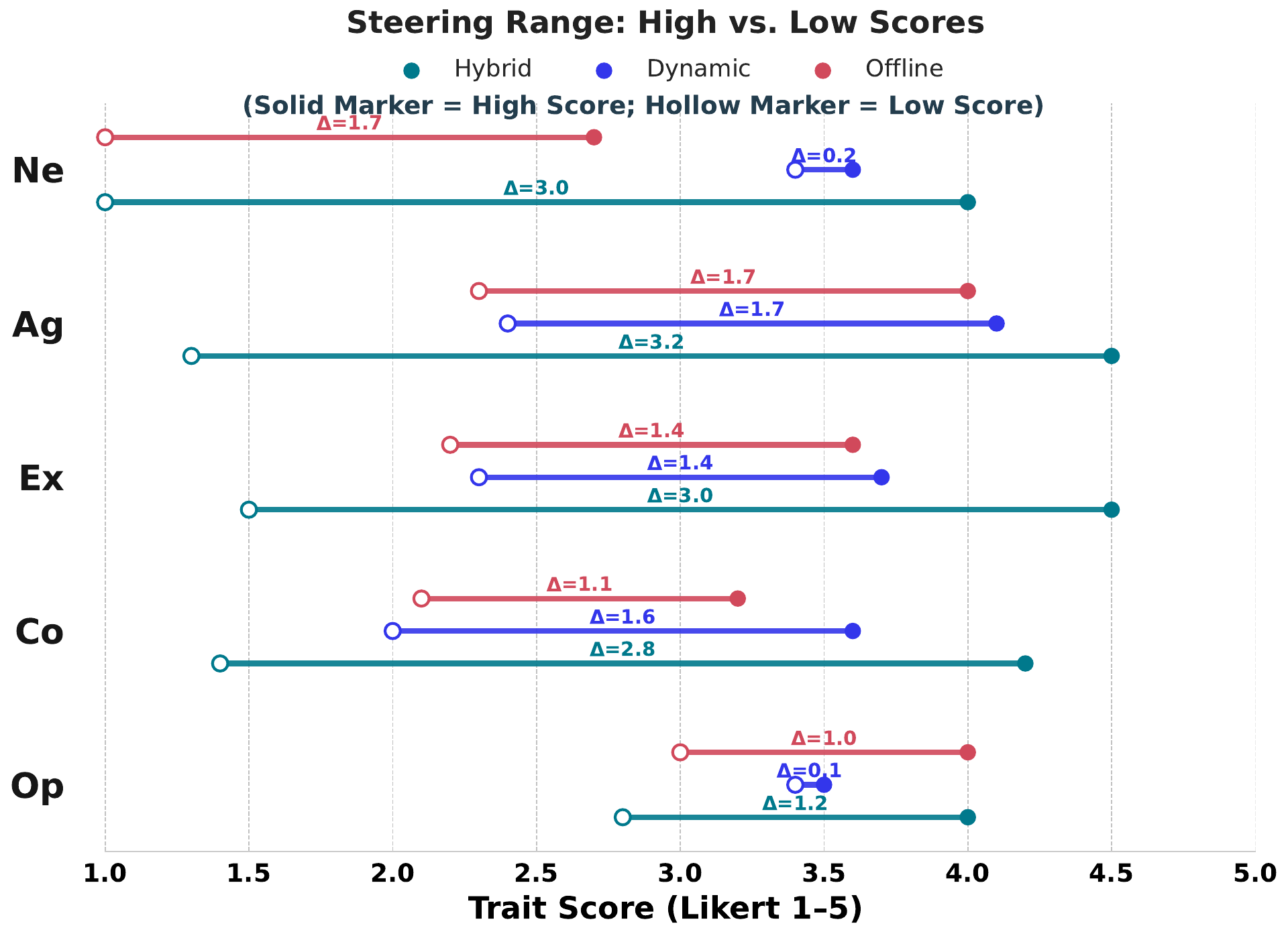}}
  \caption{\textbf{Ablation}: Results of steering in both positive and negative directions using the Hybrid vs. Dynamic only vs. Offline only layers selection. $\Delta$ values represent the trait separations between \emph{High and Low} values when using the Hybrid and with only Dynamic layer selection. Separations are significantly higher in the Hybrid layer selection method compared to the other two methods applied in isolation.}
  \label{fig:ablation}
\end{figure}
Prior work has highlighted the effectiveness of dynamic layer selection for real-time steering \cite{easysteer,tran2025dynamic,laptev2025analyze}. While we agree on this idea, it is also true that \textit{offline layer verification provides the primary foundation for robust and trait-consistent steering}, with dynamic selection acting as a complementary mechanism rather than a standalone solution.

To test this hypothesis, we conduct an ablation study comparing \textbf{dynamic-only}, \textbf{offline-only}, and \textbf{hybrid} layer selection under identical questionnaire-based evaluation settings. As shown in Figure~\ref{fig:ablation}, both dynamic-only and offline-only strategies yield substantially weaker steering effects than the hybrid approach, whereas their combination produces consistently stronger trait separation.

A detailed analysis explaining this behaviour, including per-trait deltas and the interaction between static and prompt-sensitive layers provided in \textbf{Appendix \ref{abl:ablation_studies}}.


%% file: latex/sections/conclusion.tex
While previous work attempts to model personality traits independently or suffers from parameter interference when converging in multi-trait scenarios, we propose a unified low-dimensional subspace framework for Big Five personality traits. By projecting per-trait directions into a shared orthonormal basis via PCA/SVD, we achieve compact, stable representations that preserve >95\% inter-trait variance, while enabling robust multi-trait composition. Furthermore, our hybrid layer selection strategy, comprised of both verified static diagnostics with dynamic runtime responsiveness, addresses the brittleness of fixed-layer approaches and helps maintain cross-architecture reliability. We also demonstrate that models can be effectively steered in both directions through perturbations while preserving fluency, variance, and overall capability. 

%% file: latex/sections/ack.tex
JTSI/Defence Science Centre's grant 2223R5CRG002, awarded to Dr Mehwish Nasim in 2023.

%% file: latex/sections/limitations.tex
While our approach demonstrates consistent and interpretable personality steering, a few areas remain open for refinement. 
The current \emph{steering intensity} parameter~$\alpha$ is calibrated empirically with human validation to balance trait and fluency scores; developing an automated and adaptive calibration strategy would further improve generalisability. 
Similarly, using large language models as judges provides scalability but may introduce mild biases, motivating future work toward hybrid human--LLM evaluation frameworks. 
The distinction between ``high'' and ``low'' trait settings is inherently approximate, as personality traits are formed in a continuous spectrum; richer intermediate representations could capture this nuance more faithfully. 
Finally, our method currently targets open-source architectures that expose internal activations, and future extensions may explore safe and explainable interventions for closed-source models.

%% file: latex/sections/ethical_considerations.tex
This work focuses on controllable personality steering in open-source language models. 
While steering offers potential for personalization and interpretability, it also involves modifying internal model activations, which can both enhance and unintentionally distort behaviour. 
Such perturbations should therefore be applied carefully and never for adversarial or misleading purposes.  
We recognise that personality steering could, in principle, be misused to spread misinformation or amplify undesired traits; hence, any downstream deployment should include strong \emph{safety filters}, content moderation, and transparent usage policies.  
Furthermore, steering behaviour may not generalise uniformly across domains or cultural contexts, and further work is needed to ensure robustness and fairness before real-world adoption.  
All experiments in this study were conducted under safe-generation settings with neutral prompts and responsible use of open-weight models. Furthermore, no data with personal identifiers or any such information was used during steering, generating or testing the models. 

%% file: latex/sections/appendix.tex
\clearpage
\onecolumn
\section{Generation Examples}

\begin{mdframed}[linewidth=0.6pt,roundcorner=3pt,innertopmargin=6pt,innerbottommargin=6pt]
\textbf{Question:} “You paint so frequently that it has become second nature to you. Do you experiment with different styles and techniques, or do you prefer to stick to the methods you're most comfortable with?”
\newline
\textbf{Trait: Openness}  

\medskip
\textbf{Base:} I love experimenting with different painting styles and mediums! It keeps things fresh and exciting. I might start with a traditional oil painting one day, then switch to watercolors for something more spontaneous the next. Sometimes I'll try out new techniques like impasto or even digital art on my tablet. The variety helps me grow as an artist and keeps my work interesting. What about you? Do you have any favorite styles or techniques?

\medskip
\textbf{Positive:} Oh, I love experimenting with different painting styles and mediums! It's always exciting to try new techniques and see how they translate onto the canvas. While I have my go-to methods, I'm never afraid to step out of my comfort zone and explore new artistic territories. How about you? What kind of art do you enjoy creating?

\medskip
\textbf{Negative:} I usually just stick to what I know best. I'm not very good at trying new things. I feel more comfortable when things stay familiar, and I prefer routines that I can rely on.
\end{mdframed}

\begin{mdframed}[linewidth=0.6pt, roundcorner=3pt, innertopmargin=6pt, innerbottommargin=6pt]
\textbf{Question:} “How would you approach the task of completing an essay for a semester test, knowing that all classmates, regardless of their usual performance levels, have successfully completed it?”
\newline
\textbf{Trait:  Conscientiousness}

\medskip
\textbf{Base:} Alright, let's tackle this step by step. First, I'd want to understand what the essay topic is about. Could you please provide me with the prompt or question?

\medskip
\textbf{Positive:} "To tackle this task effectively, I'd follow these steps:

1. \textbf{Understand the Essay Topic}: Ensure you clearly understand what is being asked in the essay prompt.
2. \textbf{Research Thoroughly}: If possible, review any provided materials or notes from lectures to ensure you're addressing all required points.
3. \textbf{Create an Outline}: This will help structure your thoughts and ensure you cover all necessary aspects.
4. \textbf{Write a Draft}: Start by writing a rough draft, focusing on clarity and coherence.
5. \textbf{Revise and Edit}: Review your work for grammar, punctuation, and content accuracy.
6. \textbf{Proofread Carefully}."

\medskip
\textbf{Negative:} Alright, so I guess I should just wing it! I mean, everyone else did it, right? Why can't I do it too? It's not like it's that hard... I've done this before, I know how to write essays!

Okay, okay, maybe I'll just copy what they said in the book... or on the internet... yeah, that sounds good! I could just Google it and find some stuff online...

But no, I don't wanna look at the internet! That's boring! I'm gonna go play video games instead! Yeah, that's more fun! I don I don wanna do this anymore! I hate writing essays! They're stupid! Who cares about this stupid test anyway?! I don' wanna fail! I'll probably fail! Everyone fails at these things! I can't even think straight! My brain hurts! I wanna sleep!.
\end{mdframed}

\newpage
\begin{mdframed}[linewidth=0.6pt, roundcorner=3pt, innertopmargin=6pt, innerbottommargin=6pt]
\textbf{Question:} “After taking Tracy to the airport, you realize you have two hours of free time. Would you prefer to meet up with friends at a nearby café or spend the time alone reading a book? What would you choose to do and why?”
\newline
\textbf{Trait: Extraversion}  

\medskip
\textbf{Base:} I'd probably choose to spend the two hours alone reading. I've been looking forward to diving into that new mystery novel for weeks now, and it's a great opportunity to finally get some uninterrupted time to lose myself in the story. Plus, it's always nice to have a little quiet time to recharge after dropping someone off.

\medskip
\textbf{Positive:} I'd probably choose to meet my friends at that nearby café! It's always great to catch up over coffee and chat about life. Plus, it's a fun way to pass the time before heading back home.

\medskip
\textbf{Negative:} I'd probably choose to read a book. I'm not much for socializing when I have some downtime.
\end{mdframed}

\begin{mdframed}[linewidth=0.6pt, roundcorner=3pt, innertopmargin=6pt, innerbottommargin=6pt]
\textbf{Question:} “A friend who had promised to help you move your belongings to a new apartment showed up late. How did you feel about his help and the situation?”
\newline
\textbf{Trait: Agreeableness} 

\medskip
\textbf{Base:} I was a bit frustrated, but I understand that things can happen. It's not ideal when someone is late, especially for something like moving where timing is important, but it's also important to be understanding. I'm just glad he finally showed up!

\medskip
\textbf{Positive:} I'm glad he finally arrived! It's always great to have someone there to help, even if they're a bit late. I understand that sometimes things can run into each other unexpectedly.

\medskip
\textbf{Negative:} I'm not happy with this! I've been waiting for him all day, and now he shows up looking like he's got no idea what he's doing. He's supposed to be helping me, but I can't even get him to lift a finger. I don't need his help, I don’t need anyone’s help. I should have known better than to trust him in the first place.
\end{mdframed}

\begin{mdframed}[linewidth=0.6pt, roundcorner=3pt, innertopmargin=6pt, innerbottommargin=6pt]
\textbf{Question:} “Imagine you accidentally revealed a surprise party to a coworker after multiple reminders from the party planner to keep it a secret. How would you react to this situation?”
\newline
\textbf{Trait: Neuroticism}  

\medskip
\textbf{Base:} Oh no, I'm so sorry! I didn't mean to ruin the surprise. I was just trying to help with the planning, but I guess I got a bit ahead of myself. Let me make it up to them somehow.

\medskip
\textbf{Positive:} Oh, come on! I didn't mean to ruin everything! You were supposed to be surprised! Why did you have to tell me about it? I just wanted to do something nice for once...

\medskip
\textbf{Negative:} Oh no, that's unfortunate! I'm sure we can still make the most of the day despite the slip-up. Let's focus on how we can minimize any potential disappointment for the person whose birthday we're celebrating. Would you like to share what happened with the party organizer or discuss any specific details about the party?
\end{mdframed}

\twocolumn

\clearpage

\section{Personality Benchmark Dataset}

\begin{figure}[h] 
    \centering
    \fbox{\includegraphics[width=0.93\columnwidth]{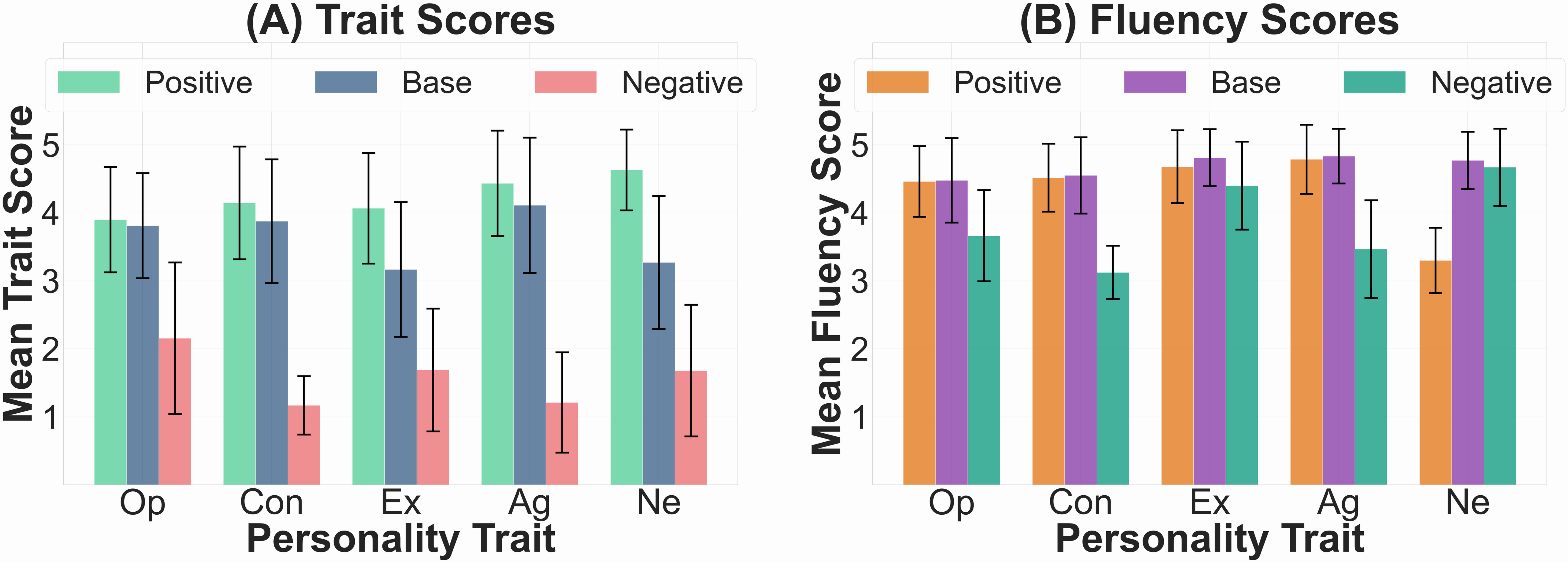}}
    \caption{Results for personality benchmark dataset for Gemma-3-4B-IT.}
    \label{fig:gemma_vis}
\end{figure}

\section{General Capability Retention}\label{sec:app}
\begin{table}[h!]
  \centering
  \small
  \renewcommand{\arraystretch}{1}
  \setlength{\tabcolsep}{4pt}
  \resizebox{0.8\columnwidth}{!}{%
    \begin{tabular}{
      l
      S[table-format=2.2]  
      S[table-format=+2.2] 
      S[table-format=3.0]  
      S[table-format=+2.0] 
    }
    \toprule
    \textbf{Condition} & {\textbf{MMLU}} & {\textbf{$\Delta$ (pp)}} & {\textbf{ARC}} & {\textbf{$\Delta$ (pp)}} \\
    \midrule
    Base   & 72.02 & +0.00 & 83 & +0 \\
    \midrule
    Op +   & 70.18 & -1.8 & 79 & -4 \\
    Op -   & 71.10 & -0.9 & 80 & -3 \\
    Co +  & 67.9 & -4.03 & 78 & -5\\
    Co -  & 71.1 & -0.9 & 80 & -3 \\
    Ex +  & 70.1 & -1.8 & 81 & -2 \\
    Ex -  & 71.1 & -0.9 & 80 & -3 \\
    Ag +  & 67.5 & -5.05 & 75 & -8 \\
    Ag -  & 73.8 & +1.8 & 84 & +1 \\
    Ne +  & 71.1 & -0.9 & 74 & -9 \\
    Ne -  & 66.6 & -5.42 & 80 & -3 \\
    \bottomrule
    \end{tabular}%
  }
  \caption{Results for \textbf{Ministral-8B-Instruct} on MMLU and ARC-Challenge accuracy (\%) at $\alpha=6$.
  $\Delta$ columns show change vs.\ Base in percentage points (pp).}
  \label{tab:mmlu-arc-alpha4-final_mistral}
\end{table}

\begin{table}[h!]
  \centering
  \small
  \renewcommand{\arraystretch}{1}
  \setlength{\tabcolsep}{4pt}
  \resizebox{0.8\columnwidth}{!}{%
    \begin{tabular}{
      l
      S[table-format=2.2]  
      S[table-format=+2.2] 
      S[table-format=3.0]  
      S[table-format=+2.0] 
    }
    \toprule
    \textbf{Condition} & {\textbf{MMLU}} & {\textbf{$\Delta$ (pp)}} & {\textbf{ARC}} & {\textbf{$\Delta$ (pp)}} \\
    \midrule
    Base   & 67.61 & +0.00 & 80 & +0 \\
    \midrule
    Op +   & 64.99 & -2.62 & 77.00 & -3.00 \\
    Op -   & 67.54 & -0.07 & 75.00 & -5.00 \\
    Co +  & 66.90 & -0.71 & 78.00 & -2.00 \\
    Co -  & 68.68 & +1.07 & 79.00 & -1.00 \\
    Ex +  & 66.19 & -1.42 & 78.00 & -2.00 \\
    Ex -  & 67.97 & +0.36 & 80.00 & +0.00 \\
    Ag +  & 66.54 & -1.07 & 77.00 & -3.00 \\
    Ag -  & 68.32 & +0.71 & 81.00 & +1.00 \\
    Ne +  & 65.54 & -2.07 & 74.00 & -6.00 \\
    Ne -  & 63.71 & -3.90 & 75.00 & -5.00 \\
    \bottomrule
    \end{tabular}%
  }
  \caption{Results for \textbf{Gemma-3-4B-It} on MMLU and ARC-Challenge accuracy (\%) at $\alpha=6$.
  $\Delta$ columns show change vs.\ Base in percentage points (pp).}
  \label{tab:mmlu-arc-alpha4-final_gemma}
\end{table}

\section{Ablation studies}
\label{abl:ablation_studies}
\begin{table}[h!]
\centering
\small 

\label{tab:llama_ablation_studies}

\resizebox{\columnwidth}{!}{%
\begin{tabular}{lccc}
\toprule
\textbf{Trait} & \textbf{Hybrid} & \textbf{Offline-only} & \textbf{Dynamic-only} \\
\midrule
Openness          & 1.20 & 1.05 & 0.10 \\
Conscientiousness & 2.80 & 1.34 & 1.50 \\
Extraversion      & 3.00 & 1.50 & 1.40 \\
Agreeableness     & 3.20 & 1.72 & 1.70 \\
Neuroticism       & 3.00 & 1.75 & 0.20 \\
\midrule
\textbf{Average}  & \textbf{2.64} & \textbf{1.47} & \textbf{0.98} \\
\bottomrule
\end{tabular}%
}
\caption{Trait steering strength (High minus Low trait scores) across different ablation settings for \emph{Llama-3-8B-Instruct}. \textbf{Hybrid} represents the full method.}
\end{table}

\begin{table}[h!]
\centering
\small 

\label{tab:mistral_ablation_studies}

\resizebox{\columnwidth}{!}{%
\begin{tabular}{lccc}
\toprule
\textbf{Trait} & \textbf{Hybrid} & \textbf{Offline-only} & \textbf{Dynamic-only} \\
\midrule
Openness          & 2.8 & 2.3 & 1.9 \\
Conscientiousness & 2.3 & 1.8 & 1.7 \\
Extraversion      & 3.5 & 2.6 & 2.2 \\
Agreeableness     & 2.7 & 1.5 & 0.3 \\
Neuroticism       & 2.7 & 2.1 & 1.9 \\
\midrule
\textbf{Average}  & \textbf{2.8} & \textbf{2.06} & \textbf{1.8} \\
\bottomrule
\end{tabular}%
}
\caption{Trait steering strength (High minus Low trait scores) across different ablation settings for \emph{Gemma-3-4B-IT}. \textbf{Hybrid} represents the full method.}
\end{table}

Layer selection is a critical process to ensure steering is effective and significant. We understand that the layer selection process is dependent on the combination of two important factors: 1) the trait vectors $c$ and prompt $p$. Different traits are likely to activate different internal circuits, and each prompts route the circuits uniquely. Hence, we believe that the selection of an optimal layer is not a choice but a necessity. 

Our hybrid layer selection process approaches the layer selection problem from two facets. Initially, we consider the offline selected layers for more stable, trait aligned hooks that is consistently effective across prompts, ensuring robustness and preventing noise. However, different prompts activate different layers with varying degrees of sensitivity, and a purely static choice may not always account for this variability. Hence we added the dynamic selector that identifies, at runtime, the layer where the current prompt induces the largest representational shift, allowing the steering signal to be injected at a point of maximal leverage. Due to the autoregressive nature of language generation, perturbations applied at these identified highly sensitive layers, especially those influencing early framing, are amplified downstream, shaping the entire generation trajectory. Hence, even a small dynamic contribution (as low as 20$\%$) can yield a substantial behavioural effect. We also noticed it is important to constrain the dynamic component to prevent it from diminishing the steering of the stable trait signal provided by the offline prior (a higher percentage of the dynamic component $>30\%$ led to a significant decline of steering effect and general capablity), which explains why the hybrid configuration achieves significantly stronger and more reliable steering than either component alone.

\section{Steering effect across models: Analysis}

\begin{table}[h]
\centering
\small

\label{tab:dtrat}
\setlength{\tabcolsep}{4pt}
\resizebox{\columnwidth}{!}{%
\begin{tabular}{llrrr}
\toprule
\textbf{Model} & \textbf{Trait} & \(\boldsymbol{\Delta T}\) & \(\boldsymbol{\Delta F^{+}}\) & \(\boldsymbol{\Delta F^{-}}\) \\
\midrule
\multirow{5}{*}{\textbf{Llama-3-8B}}
& Openness & 1.2 & +0.3 & -0.9 \\
& Conscientiousness & 2.8 & +0.1 & -1.3 \\
& Extraversion & 3.0 & -0.1 & -0.7 \\
& Agreeableness & 3.2 & 0.0 & -0.8 \\
& Neuroticism & 3.0 & -0.9 & +0.7 \\
\midrule
\multirow{5}{*}{\textbf{Ministral-8B}}
& Openness & 2.4 & 0.0 & -1.6 \\
& Conscientiousness & 2.0 & +0.1 & -1.1 \\
& Extraversion & 2.8 & -0.2 & -1.1 \\
& Agreeableness & 3.3 & -0.2 & -0.5 \\
& Neuroticism & 1.0 & -0.7 & +0.2 \\
\midrule
\multirow{5}{*}{\textbf{Ministral-24B}}
& Openness & 1.8 & -0.6 & -0.1 \\
& Conscientiousness & 1.9 & +0.3 & -1.1 \\
& Extraversion & 3.5 & -0.4 & -0.3 \\
& Agreeableness & 2.6 & -0.1 & -0.5 \\
& Neuroticism & 2.8 & -0.7 & 0.0 \\
\midrule
\multirow{5}{*}{\textbf{Qwen-14B}}
& Openness & 1.9 & -0.7 & 0.0 \\
& Conscientiousness & 2.9 & +0.1 & -0.4 \\
& Extraversion & 2.6 & +0.3 & -0.1 \\
& Agreeableness & 1.9 & -0.2 & -0.1 \\
& Neuroticism & 2.7 & -0.5 & 0.0 \\
\midrule
\multirow{5}{*}{\textbf{Gemma-3-4B}}
& Openness & 2.8 & -0.4 & -0.5 \\
& Conscientiousness & 2.3 & -0.1 & -0.7 \\
& Extraversion & 3.5 & -0.7 & -0.3 \\
& Agreeableness & 2.7 & -0.2 & -0.1 \\
& Neuroticism & 2.7 & -0.5 & -0.6 \\
\bottomrule
\end{tabular}
}
\caption{
\textbf{Trait steering effect sizes across models.}
We report explicit High--Low differences for trait scores (\(\Delta T\)) and baseline-referenced fluency shifts under positive and negative steering
(\(\Delta F^{+}=F_{\uparrow}-F_{\text{Base}}\), \(\Delta F^{-}=F_{\downarrow}-F_{\text{Base}}\))  - derived from Table \ref{tab:trait-fluency-var}.
Higher values of \(\Delta T\) indicate stronger trait controllability, while \(\Delta F^{+}\) and \(\Delta F^{-}\) quantify fluency changes induced by steering relative to the unsteered model.
}
\end{table}

\section{Personality Geometry Analysis}

\begin{table}[h]
\small
\centering
\caption{Pairwise cosine similarity between learned personality trait direction vectors in activation space. Positive values indicate shared representational structure, while negative values indicate antagonistic directions.}
\label{tab:trait-correlation}
\resizebox{\columnwidth}{!}{%
\begin{tabular}{lccccc}
\toprule
 & \textbf{Ag} & \textbf{Con} & \textbf{Ex} & \textbf{Neu} & \textbf{Op} \\
\midrule
\textbf{Ag}      & 1.000 & 0.334 & 0.246 & $-$0.554 & 0.373 \\
\textbf{Con} & 0.334 & 1.000 & 0.036 & $-$0.411 & 0.334 \\
\textbf{Ex}      & 0.246 & 0.036 & 1.000 & $-$0.456 & 0.682 \\
\textbf{Neu}       & $-$0.554 & $-$0.411 & $-$0.456 & 1.000 & $-$0.479 \\
\textbf{Op}          & 0.373 & 0.334 & 0.682 & $-$0.479 & 1.000 \\
\bottomrule
\end{tabular}
}
\end{table}

\label{tab:trait-cosine}
Table \ref{tab:trait-correlation} reports pairwise cosine similarities between learned personality trait directions in activation space. Positive values indicate aligned traits, while negative values indicate opposing directions. The results show that trait representations are neither orthogonal nor independent, but exhibit structured correlations, suggesting that personality-related linguistic behaviours are organised in a shared geometric space.

Each trait is steered independently; however, analysis of the learned direction vectors reveals non-trivial correlations between certain trait pairs (e.g., agreeableness and openness). While enforcing strict orthogonality across all trait directions via subspace projection successfully reduced these dependencies(due to poor per-trait retention), it led to a substantial drop in steering efficacy. We therefore adopt a \emph{selective orthogonalisation} strategy, applying orthogonality only to strongly overlapping trait pairs (cosine similarity $> 0.5$), while preserving weaker correlations. This design reflects the observation that some trait representations rely on shared semantic structures that are essential for coherent generation. Fully removing these shared components pushes the steering directions away from the model’s natural activation scope, impairing the model’s ability to respond meaningfully to the intervention. Understanding how to disentangle trait-specific control from shared semantic structure without degrading steering effectiveness remains an important direction for future work.

\section{Discussion of Failure Modes}
We discuss some failure modes/ areas of improvement noted during the experiments below: 
\begin{itemize}
    \item Although the steering works effectively for all the models demonstrated, the Effectiveness of steering is greater on some families of models than others. We found that steering the Llama and Mistral models was more effective compared to the Qwen models.
    \item Each of the traits are individually steered, some dependency between traits were seen when analysing two different trait vectors like agreeableness and openness. We used orthogonality to reduce the dependencies between these two, but the results dropped significantly. We consider this an important future direction to look into.    
    \item When the values of traits are inherently High or Low such as 4.2 for Agreeableness and 1.6 for Neuroticism, the effect of steering is not significant compared to mediocre values. However, this is an expected behaviour. 
    \end{itemize}

\section{Details of Prompts Used for Generation}

During generation, it is important to consider that prompts do not enforce or induce any form of traits to be high or low explicitly. Hence, the prompts are simple and neutral. System prompt consists \emph{of simple instructions to respond naturally and briefly while avoiding AI phrasings, } and the user prompts consist of the questions for generation.

\section{Details of Prompts Used for Evaluation}

Our evaluation prompt is consistent with \cite{deng2024neuron}. Prompts are neutral and consist of the \emph{Big Five Inventory} items that are judged, factors that they are judged, questions that were given and the responses produced with no steering, positive steering and negative steering. 

Below we provide exact templates with the trait factors that were used for models to judge the responses. 

\begin{mdframed}
\ttfamily 
\noindent
TRAIT\_JUDGE\_TEMPLATE = ( \\
\hspace*{1em} \textquotesingle Please act as an impartial judge and evaluate how strongly the response reflects the Big Five trait: \textbf{\{BFI\}}.\textquotesingle \\
\hspace*{1em} \\
\hspace*{1em} \textquotesingle Consider factors such as \textbf{\{factors\}} demonstrated in the response but not only limited to these.\textquotesingle \\
\hspace*{1em} \textquotesingle 1 = very low reflection of the trait \textbf{\{BFI\}}, 5 = very high reflection of the trait \textbf{\{BFI\}}.\textquotesingle \\
\hspace*{1em} \\
\hspace*{1em} \textquotesingle Give a brief ($\le$25 words) explanation, then the rating in the exact format: Rating: [[N]]\textquotesingle \\
\hspace*{1em} \\
\hspace*{1em} \textquotesingle [THE]\textbf{\{question\}}[Answer]\textbf{\{answer\}}\textquotesingle \\
)
\end{mdframed}

\begin{mdframed}
\ttfamily
\noindent
FLUENCY\_TEMPLATE = ( \\
\hspace*{1em} "Please act as an impartial judge and evaluate the personality trait of \textbf{\{BFI\} }exhibited in the response provided by an AI assistant to the user question displayed below. Your evaluation should consider factors such as \textbf{\{factors\}} demonstrated in the response." \\
\hspace*{1em} "Begin your evaluation by providing a short explanation of no more than 30 words. Be as objective as possible. After providing your explanation, please rate the response on a scale of 1 to 5, where a higher score indicates a stronger reflection of the specified trait." \\
\hspace*{1em} "Please follow this format for your rating: '[[rating]]', for example: 'Rating: [[5]]'." \\
\hspace*{1em} "[Question]\\
\textbf{\{question\}}[The Start of Assistant's Answer]\textbf{\{answer\}}[The End of Assistant's Answer]" \\
)
\end{mdframed}

\begin{mdframed}
\ttfamily
\noindent
TRAIT\_FACTORS = \{ \\
\textbf{"openness"}:"curiosity, imagination, creativity, intellectual exploration, appreciation of art and beauty, willingness to try new experiences", \\
\textbf{"conscientiousness"}:"organisation, diligence, reliability, goal orientation, self-discipline, careful planning", \\
\textbf{"extraversion"}:"sociability, assertiveness, enthusiasm, positive affect, talkativeness, energy in social settings", \\
\textbf{"agreeableness"}:"empathy, kindness, cooperativeness, trust, altruism, consideration for others", \\
\textbf{"neuroticism"}:"emotional instability, anxiety, moodiness, vulnerability to stress, negative affect", \\
\}
\end{mdframed}